# Foundations for Digital Twins

Finn Wilson [1,2], Regina Hurley [1,2], Dan Maxwell [2,3], Jon McLellan [2,3], and John Beverley [1,2,4,*]

[1] *Department of Philosophy, University at Buffalo, Buffalo, NY, USA*
[2] *National Center for Ontological Research, University at Buffalo, Buffalo, NY, USA*
[3] *KadSci, Fairfax, Virginia, USA*
[4] *Institute for Artificial Intelligence and Data Science, University at Buffalo, Buffalo, NY, USA*

**Abstract**

The growing reliance on digital twins across industry brings with it interoperability challenges. Ontologies are a well-known strategy for addressing such challenges, though given the complexity of digital twins there are risks of ontologies reintroducing interoperability issues. To avoid such pitfalls, we defend characterizations of digital twins within the context of the Common Core Ontologies. We provide definitions and a design pattern relevant to the domain, and in doing so a foundation on which to build more sophisticated ontological content related and connected to digital twins.

## 1. Introduction

The concept of digital twins was first introduced by NASA in the 1960s as part of the Apollo 13 program [1]; decades would pass before the first documented definition was offered in 2003 [2]. Digital twins were envisioned to be sophisticated virtual representations of physical systems used to track, evaluate, and assess those systems through real-time updates. Characterizations of digital twins evolved considerably over subsequent decades. The advent of the Internet of Things (IoT) brought a need for efficient, secure, interactions across interconnected devices and software systems [3]; researchers accordingly observed the value of digital twins for physical assets and manufactured goods [4], as well as manufacturing processes [5], business logic [6], and the environment [7]. Digital twins contribute to sophisticated integrations of technologies, frameworks, products, and so on. Indeed, the global digital twin market is expected to top 73 billion USD by 2027, with companies such as Meta and Nvidia capitalizing on this technology [8].

As with any data-driven endeavor, the specter of semantic interoperability looms over digital twins. A 2020 report by The National Institute for Standards and Technology (NIST) estimated costs emerging from the lack of interoperability across industrial datasets as between 21-43 billion USD [9]. Leveraging digital twins in this environment runs the risk of exacerbating interoperability costs. On the one hand, ambiguity over what counts as a "digital twin" results in what we might call *social interoperability challenges* [10, 11]. On the other hand, differing data formats, coding standards, and jargon result in well-known *technical interoperability challenges* [12]. Symptomatic of each is the presence of *data silos* [13], datasets representing nearby domains that cannot be easily integrated using standard computing techniques. Because digital twins rely on the integration and synthesis of real-time data from disparate sources, data silos are particularly problematic. Achieving meaningful digital twin data exchange requires overcoming hurdles that underwrite silos.

Ontologies – controlled vocabularies of terms and logical relationships among them – are a well-known resource for addressing semantic interoperability challenges [14]. Ontologies have been leveraged to support data standardization, integration, machine learning, natural language processing, and automated reasoning [15] in fields such as biology and medicine [16] and proprietary artificial intelligence products, such as Watson [17]. Digital twin IoT researchers are well-aware of the benefits of ontologies [18, 19] as evidenced by the World Avatar digital twin project [20] among others [21]. If pursued without oversight, however, combining digital twins and ontologies can easily recreate semantic interoperability problems [22]. This occurs, for instance, when ontologies representing content specific to digital twins are created without reflection on how they might integrate with nearby ontologies, i.e. *ontology silos*.

Decades ago, recognition of such undesirable consequences led to the creation of ontology 'foundry' efforts [23, 24] aimed at creating ontologies in accordance with common standards. Among the principles underwriting most such foundry efforts is that ontologies should extend from a common top-level architecture: Basic Formal Ontology (BFO) [25], a highly-general ontology designed to contain classes and relations representing phenomena common to all areas of the world, e.g. *object*, *process*, *part of*.[1] BFO is designed to be extended to more specific domains, and as such is used in over 600 ontology initiatives, providing a rich ecosystem covering areas such as biomedicine, manufacturing, defense and intelligence, and education, to name a few. We maintain the best strategy for leveraging ontologies to address semantic interoperability challenges arising from digital twins will be one that leverages BFO. To that end, in what follows we explore common definitions of "digital twin" and identify themes and issues with the goal of constructing a BFO-based ontologically precise definition for this expression and nearby phenomena. We employ an extension of BFO – the Common Core Ontologies (CCO) [26] suite – as a foundation on which to construct our definitions, with a particular emphasis on information design patterns characteristic of the suite.[2] In doing so, we provide a firm ontological foundation on which to construct more sophisticated representations of digital twins within the BFO ecosystem.

## 2. Related Work

There are numerous ontological characterizations of digital twins [27, 28, 29, 30]; most do not leverage a top-level ontology, and so run the risk of creating ontology silos. Nevertheless, ontological characterizations leveraging a top-level do exist, e.g. the ISO digital twins in manufacturing standard [31] has a corresponding BFO-conformant ontology [32]. Whereas this digital twin ontology is specific to manufacturing, our proposal characterizes digital twins more broadly. Another example characterizes basic requirements for an ontology of digital twins under the scope of the Unified Foundational Ontology (UFO) and provides a set of competency questions for evaluation [33]. We stick with BFO owing to its wide use but address several competency questions identified in this work. For example, our characterization of digital twins reflects levels of granularity, relations among digital twin types, and digital twin updates from physical assets.

---

[1] BFO is under CC BY 4.0: https://github.com/BFO-ontology/BFO-2020
[2] CCO under the BSD-3: https://github.com/CommonCoreOntology/CommonCoreOntologies

## 2.1 Definitions of "Digital Twin"

Exploring the range of "digital twin" definitions reveals common themes and limitations [34]. **Table 1** displays 11 sample definitions, several of which are frequently cited in discussions of digital twins. Inspired by these definitions, we provide a preliminary definition of "digital twin", which we leverage here to highlight gaps in the definitions of **Table 1**: *A virtual representation designed to either represent updates of and send updates to a physical asset or provide a model for how such a physical asset can be created*. The subsequent section shows how to represent this characterization in the BFO ecosystem. Before turning there, we here evaluate definitions in **Table 1**.

One theme is the treatment of digital twins is as virtual representations designed to represent some physical asset or system; another is that they be designed for synchronization with represented assets. While important, defining "digital twin" as requiring such interaction excludes digital twins that have been created in, say, anticipation of the manufacturing of the corresponding physical asset. However, digital twin "prototypes" may be created as blueprints for physical assets they will ultimately represent [35]. Definitions B, C, D, E, G, H, and I in **Table 1** problematically require a corresponding physical asset for something to count as a digital twin, indicated by an "X" in the "SYN" column.

Definitions differ with respect to scope, some being narrower than others [10]. For example, the restriction to physical manufactured products in definition A excludes digital twins of human bodies [36] and Earth [7], among other natural entities. Similar remarks apply to definitions B ("as-built vehicle"), G ("physical production lines"), and I ("manufacturing element"). Definition A is, moreover, too exclusive in another sense, as it requires digital twins "fully" describe a physical asset across levels of granularity; no digital twin can be so complete. Similar remarks apply to definition F with respect to "comprehensive" descriptions. The "SCP" column reflects definition scope problems.

Digital twins are often conflated with nearby entities [37, 38]. For example, digital twins are sometimes conflated with "digital shadows", the latter providing only one-way communication from a physical asset to a virtual representation. Similar remarks apply to conflation with "product avatars" [39]. Definitions B and H subsume digital twins under "simulation", though the latter are snapshots of a system state used for prediction and analysis [40], while digital twins are synchronized for real-time evaluation. Definition G treats digital twins as combinations of virtual representations and physical assets, conflating a synchronizing system and one of its parts. The "TAX" column identifies definitions exhibiting improper taxonomic characterization.

**Table 1 Definitions of "Digital Twin"**

| ID | Definition | SYN | SCP | TAX |
|---|---|---|---|---|
| A | Virtual information constructs that fully describe potential or actual physical manufactured products from the micro atomic level to the macro geometrical level [2] |  | X |  |

| B | Integrated multiphysics, multiscale, probabilistic simulation of an as-built vehicle or system that uses…physical models, sensor updates, fleet history, etc., to mirror the life of its corresponding flying twin [41] | X | X | X |
|---|---|---|---|---|
| C | Virtual representation of a physical system (and its associated environment and processes) that is updated through the exchange of information between the physical and virtual systems [42] | X | | |
| D | Digital replica of a living or non-living physical entity…to gain insight into present and future operational states of each physical twin [43] | X | | |
| E | Virtual representation of an object or system that spans its lifecycle, is updated from real-time data, and uses simulation, machine learning, and reasoning to help decision-making [44] | X | | |
| F | Comprehensive physical and functional description of a component, product, or system together with all available operational data [45] | | X | |
| G | Functional system formed by the cooperation of physical production lines with a digital copy [46] | X | X | X |
| H | A simulation based on expert knowledge and real data collected from the existing system [47] | X | | X |
| I | Fit for purpose digital representation of an observable manufacturing element with synchronization between the element and its digital representation [31] | X | X | |

## 3. Ontological Characterization of Digital Twins

The Common Core Ontologies (CCO) suite extends from BFO and inherits its methodological commitments [48], such as aiming to represent reality rather than merely concepts about it. CCO is a bridge from the highly general, rather abstract, BFO to the more specific content relevant to digital twins. We introduce relevant elements from BFO/CCO as needed.[3]

### 3.1 Digital Twins as Information

**Table 2** describes BFO/CCO definitions useful for our characterization of digital twins. Digital twins are plausibly described as *information*; in CCO terms, they fall under the class **information content entity**,[4] a subclass of the BFO class **generically dependent continuant**, where one finds entities that may be copied across bearers. To illustrate, observe that distinct computer monitors could bear 'π' or '3.14159265358979323…', each conveying the same information said to *generically depend on* the monitors, which are **information bearing entities** when they enter such a relation. Accordingly, a given digital twin might be said to *generically depend on* some computer hardware, such that if all relevant hardware ceased to exist, so would the corresponding digital twins.

**Table 2** BFO and CCO Elements Leveraged

| Label | Type | Definition |
|---|---|---|
| *continuant (BFO)* | class | An entity that persists, endures, or continues to exist through time while maintaining its identity |
| *occurrent (BFO)* | class | An entity that unfolds itself in time or is the start or end of such an entity or is a temporal or spatiotemporal region |
| *process (BFO)* | class | An occurrent p that has some temporal proper part & for some time t, p has some material entity as participant |

---

[3] An OWL version of our proposal can be found here: https://github.com/Finn1928/Digital-Twins-Ontology/tree/main
[4] In the sequel, **bold** will be used to represent classes, *italics* to represent relations.

| *x generically depends on y (BFO)* | object property | x is a generically dependent continuant & y is an independent continuant that is not a spatial region & at some time t there inheres in y a specifically dependent continuant which concretizes x at t |
|---|---|---|
| *generically dependent continuant (BFO)* | class | An entity that exists in virtue of the fact that there is at least one of what may be multiple copies which is the content or the pattern that multiple copies would share |
| *material entity (BFO)* | class | An independent continuant that has some portion of matter as continuant part |
| *information content entity (CCO)* | class | A generically dependent continuant that generically depends on some information bearing entity & stands in relation of aboutness to some entity |
| *stasis (CCO)* | class | A process in which one or more independent continuants endure in an unchanging condition |
| *environmental feature (CCO)* | class | A material entity that is either a natural or man-made feature of the environment |
| *change (CCO)* | class | A process in which some independent continuant endures & 1) one or more of the dependent entities it bears increase or decrease in intensity, 2) it begins to bear some dependent entity or 3) it ceases to bear some dependent entity |
| *descriptive ice (CCO)* | class | Information content entity that consists of a set of propositions or images that describe some entity |
| *directive ice (CCO)* | class | Information content entity that consists of a set of propositions or images that prescribe some entity |
| *representational ice (CCO)* | class | Information content entity that represents some entity |
| *information bearing entity (CCO)* | class | Object upon which an information content entity generically depends |
| *x represents y (CCO)* | object property | x is an instance of information content entity, y is an instance of entity, & z is carrier of x & x is about y in virtue of there existing an isomorphism between characteristics of z & y |
| *x describes y (CCO)* | object property | x is an instance of information content entity & y is an instance of entity & x is about the characteristics by which y can be recognized or visualized |
| *x prescribes y (CCO)* | object property | x is an instance of information content entity & y is an instance of entity & x serves as a rule or guide for y if y an occurrent, or x serves as a model for y if y is a continuant |

Digital twins often represent some existing physical asset.[5] A digital twin might, however, serve as a prototype that prescribes how to create a future physical asset. Noting this, Grieves and Vickers distinguish between Digital Twin Instance (DTI) – which describes a physical product to which a digital twin remains linked throughout the life of the product – and Digital Twin Prototype (DTP) – information needed to produce a physical product meeting the specifications of a digital twin [2]. **Figure 1** displays how we may respect this distinction by leveraging specializations of **information content entity** that are prescriptive – such as the information comprising a blueprint – or representational – such as the content of a photograph. DTIs are plausibly understood as at least representational, and so falling under **representational information content entity** in CCO. **Representational information content entities** represent in a variety of ways. For example, the content of a painting of Napoleon Bonaparte represents the former emperor since the content *generically depends on* the painting which in turn bears similarities to

---

[5] That is, a **material entity** or **process** that **material entities** *participate in*.

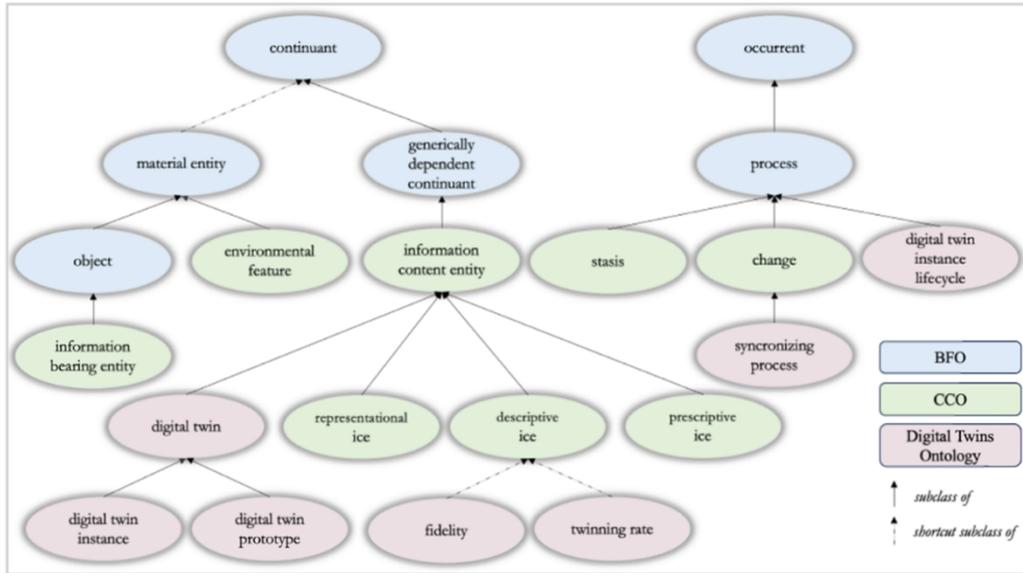

**Figure 1:** Bridging BFO, CCO, and the proposed Digital Twin Ontology

Napoleon. Similarly, a digital twin represents some physical asset insofar as it *generically depends on* computer hardware that bears similarity to that physical asset. Appeal to "isomorphism" in the definition of *represents* is understood as relative to the type of entities involved, i.e. an isomorphism for one pair of entities need not share much in common with an isomorphism between a distinct pair of entities. The arrangement of Napoleon's body parts in a painting by Jacques Louis David was meant to reflect the actual arrangement of Napoleon's body; the arrangement of components in a digital twin is not meant to reflect the arrangement of parts of the corresponding physical asset, though some manner of isomorphism between the digital twin and physical asset exists, such that were the latter to be physically altered then the digital twin might no longer represent the physical asset.

DTIs need not be solely representational. A given DTI may have parts that *describe* or *prescribe* other entities, e.g. the digital twin of Truist Park [49] includes descriptions of historical baseball players as well as directions for how to navigate the park. The digital twin both represents the park while having parts that are not merely representational.

**Table 3** Digital Twins Ontology Elements

| Label | Type | Definition |
|---|---|---|
| *digital twin* | class | An information content entity that is either represents a material entity or process relative to some granularity or prescribes an arrangement of types in such a manner as to produce a digital twin instance |
| *digital twin instance*[6] | class | A digital twin that represents some material entity or process |
| *digital twin prototype* | class | A digital twin that prescribes an arrangement of types in such a manner as to produce a digital twin instance |
| *synchronizing process* | class | A change during which a digital twin instance is updated based on real-time information transmitted from the entity it represents |
| *x is counterpart material entity y* | object property | x represents y, x is a digital twin instance, y is a material entity & x & y participate in a synchronizing process |
| *x is counterpart process y* | object property | x represents y, x is a digital twin instance, y is a process & x participates in a synchronizing process that overlaps with y |

[6] **Digital twin instance** is an OWL inferred subclass of **representational ice**.

| *twinning rate* | class | A ratio measurement content entity that is a measurement of the rate at which synchronization occurs between a digital twin instance and the entity it represents |
| --- | --- | --- |
| *fidelity* | class | A measurement information content entity that is a measurement of the number of information types, their accuracy, generality, and quality transferred between a digital twin instance and what it represents |
| *digital twin instance lifecycle* | class | A process that consists of all and only processes in which either 1) a digital twin instance and the material entity it represents participate or 2) a digital twin instance participates and the process it represents is a proper process part |

In CCO, the *represents* relation holds between instances.[7] If there is no instance for a DTP to *represent*, then that DTP cannot be a **representational information content entity**. This seems correct as DTPs seem best understood as plans or blueprints rather than as representations. In CCO, prescriptive entities of this sort fall under the class **directive information content entity**, which in every case *prescribe* some instance. Here again, however, there is no instance that a DTP can be said to *prescribe*. The issue we are encountering is not new. There are known challenges to characterizing unrealized plans and blueprints in BFO and CCO. CCO maintains an extension – the Modal Relations Ontology (MRO) [48] – developed in part to address this issue. MRO introduces the *modal object property* under which duplicates of all CCO relations fall as sub-relations. Users then model actuality using the original CCO relations and actual instances while users model possibility using the CCO relations under *modal object property* and possible instances. To apply this strategy here one would need to create an instance which the DTP possibly *prescribes*, but that suggests a misunderstanding. A DTP need not prescribe an actual or possible instance.

We maintain that a given DTP is intended to prescribe possible arrangements of classes and relationships among them. A DTP for a planned motorcycle series is not about any motorcycle instance that might emerge from production, though it does prescribe arrangements of portions of rubber and metal, properties of shape, size, and thermal conductivity, relations of parthood and dependence, and so on. This does not mean that a given DTP prescribes anything regarding some specific instance of, say, a portion of metal; there may be no such portion of metal having characteristics prescribed by the DTP. The prescription exhibited by DTPs aims at the class-level rather than instance-level.[8] This proposal would require changing CCO *prescribes*, which has range instances of the class **entity**. We believe this is warranted as our proposal more accurately reflects the intentions behind unrealized plans or blueprints than alternatives like MRO.[9]

Pursuing either path leads to DTPs counting as prescriptive entities – or **directive information content entities** – insofar as they serve as a model for the creation of an entity that would plausibly serve as a physical twin. In the event the relevant physical twin is created, DTP instance counts also as a DTI instance, i.e. a digital twin **directive information content entity** may be a **representational information content entity**. This tracks the intuition that when a physical asset is created satisfying a DTP prescription, the digital twin both prescribes and represents the physical asset.

## 3.2 Counterparts of Digital Twins

---

[7] Because CCO adopts the OWL2 direct semantics, all object properties are intended to hold between instances.

[8] Our proposal is general. "Superman" has superhuman qualities, arrangements of real classes, e.g. flight, strength, etc.

[9] Implementing this proposal requires using OWL Full, since OWL 2 with the direct semantics does not permit class-level relationships. For those who prefer practicality over accuracy, MRO remains an option.

DTIs have in every case some counterpart, for example, the real-world wind turbine represented by a wind turbine digital twin. DTIs should not be restricted to physical assets, as researchers often construct digital twins for manufacturing [46] and design processes [12]. Relevant here is that CCO adopts BFO's fundamental division between **occurrent** and **continuant**. **Occurrents** are extended over time and have temporal parts, such as eating or walking, which are examples of the **process** subclass of **occurrent**. Instances of **continuant** lack temporal parts, endure through time, and *participate in* instances of **occurrent**. CCO extends **process** with subclasses, such as natural processes, agential acts, mechanical processes, and so on, thus providing resources to distinguish physical assets from process counterparts of digital twins.

There is a need to connect digital twins, where possible, to relevant counterparts. Our strategy is to introduce sub-properties of *represents* reflecting representation, tracking, and synchronization. We introduce *is counterpart process* with range **process**. Similarly, we introduce *is counterpart material entity* since physical counterparts of DTIs plausibly fall under the BFO **continuant** subclass **material entity**, instances of which have matter as parts. CCO provides resources to draw a further distinction between **artifacts** - **material entities** designed to achieve some function - and **environmental features** - **material entities** such as rivers, wind, Earth, and so on. Our proposal thus distinguishes among the wide variety of digital twin counterparts, whether natural, manufactured, or processual. Because in BFO such entities often *participate in* **processes**, there is a line connecting digital twins representing processes to those representing physical assets *participating in* them.

## 3.3 Twinning

Digital twins are often updated with real-time information about changes in the corresponding physical counterpart, which can be accounted for in CCO using the class **change**, roughly, a **process** in which a **continuant** gains or loses one or more properties. CCO contains a rich hierarchy reflecting varieties of such gains and losses. For example, if a vehicle *participates in* an increase of its thermal energy, this amounts to a **change** in which one temperature quality of the vehicle is replaced. Gain or loss of properties is not the only way in which physical counterparts might change. A wind turbine plausibly *participates in* a change when one of its fan blades is replaced. This involves replacement of a material part of the turbine, rather than replacement of its properties. Such change can be captured by observing a change of material parts will in every case involve a change in properties. The wind turbine initially, say, had a worn blade that is later, say, replaced by a fresh blade.

Supposing a given sensor system is working correctly, a **change** in a physical counterpart will initiate a signal-sending **process**, during which a signal will be sent to and received by the corresponding digital twin. Because the digital twin is an **information content entity**, updating the digital twin requires updating the computer system on which it *generically depends*. Like the physical counterpart of the digital twin, updates to the computer system can be represented as a **change** during which properties are gained or lost. For example, suppose a decelerating vehicle is the physical counterpart of a digital twin that is updated with information regarding velocity. Circuitry within the relevant computer hardware *participate in* some **change** during which qualities of the hardware are replaced with others. The corresponding digital twin that *generically depends on* the hardware may then have updated parts, such as a **descriptive information content entity** that *describes* the velocity of the vehicle as decelerating.

**Figure 2** illustrates a digital twin instance updating to reflect a change in temperature from the ground vehicle which it represents, which involves synchronization, or the real-time updating of the digital twin instance based on changes in its counterpart. An important feature of this relationship is the so-called **twinning rate** at which real-time updates can be conducted and sustained over time. CCO provides resources for the measurement of such rates within scope of its measurement unit module.

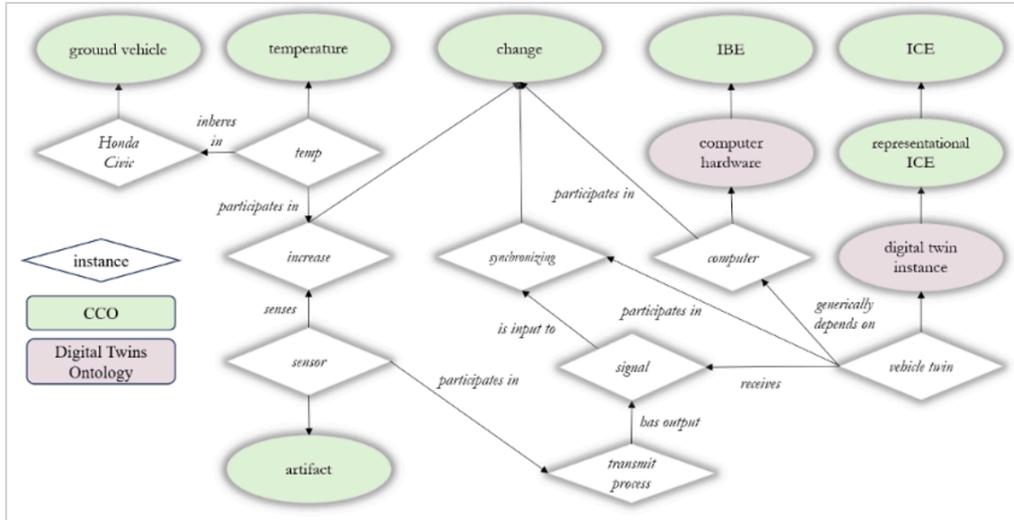

**Figure 2:** Synchronization between a ground vehicle and its digital twin instance.

### 3.4 Fidelity as Granularity Partitions

Important to digital twins is the degree of *fidelity* desirable between the virtual representation and what it represents [10, 11]. Digital twin development is often pursued iteratively, where sub-components of the twin are added or refined in response to changes in the physical counterpart; a given digital twin may change to emphasize different levels of fidelity and relationships. In either case, mereological relationships appear relevant.

We may characterize fidelity in terms of the *theory of granular partitions* [50]. A given digital twin of a vehicle may have a part representing the vehicle's engine but not other engine parts, such as pistons. We might think of this as a projection onto a whole that does not project onto all proper parts of the whole. **Figure 3** illustrates. In **Scene 2**, a partition of the vehicle and its engine might not project onto other vehicle parts, such as the front window. **Scene 3** illustrates when a material entity is added to the engine, namely, a piston, in which "the object targeted by the root cell…remains the same". Lastly, the root of the digital twin granular partition could be extended. **Scene 4** illustrates such a case where a digital twin represents more than one vehicle so "the target of the original root cell is always a proper part of the extension's root cell". Mereological relationships across granular partitions provide partition connections. A digital twin engine has a digital twin piston as part under some partition because the **material entity** counterpart of the engine does.

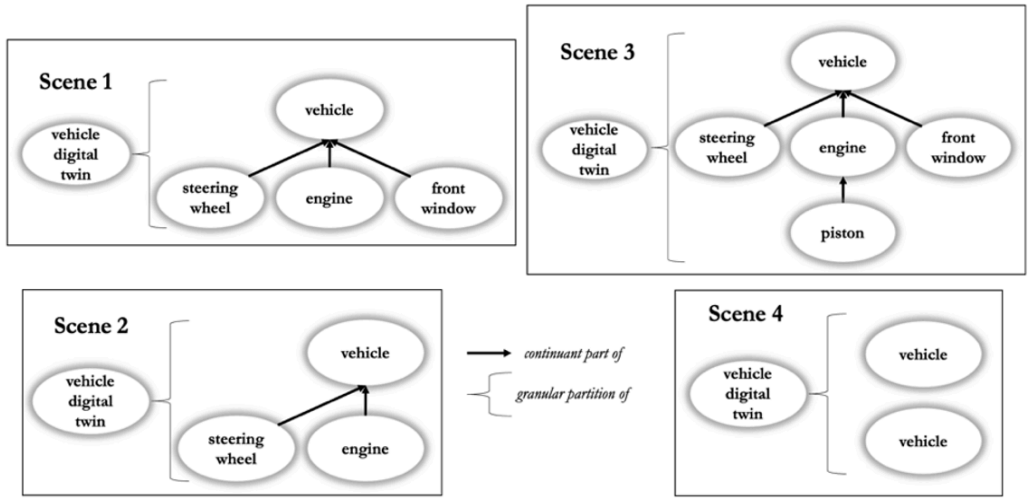

**Figure 3:** Three granular partitions of a ground vehicle.

Granular partitions provide a guide for how fidelity might change during the use of digital twins where we understand as a measurement of the types of information transferred between a digital twin instance and what it represents. This might include information regarding the digital twin counterpart's temperature, overall health, production capabilities, and so on. In each case, the degree of fidelity is relative to a granular partition of interest as contrasted with the granular partitions that are not of interest. For example, we might say the granular partition of the vehicle referenced above does not exhibit a high degree of fidelity. We should take care as fidelity cannot be reduced to the number of parts in each partition. A partition that covers, say, the transmission of information regarding the temperature and weight of an engine has a higher fidelity than a one covering only temperature. This raises no special modeling problem, however. Just as, according to our ontological design patterns, the engine would be part of the vehicle, we can say that parts of the vehicle bear qualities such as temperature and weight. Moreover, different granular partitions will contain material entities that bear different qualities, much like different granular partitions contain material entities having different parts.

## 4. Conclusion

Our goal has been to avoid interoperability pitfalls by characterizing digital twins within BFO and CCO. We envision this work to be foundational for more sophisticated ontological representations of digital twins within the BFO ecosystem. Moreover, we envision our work will be extendable characterizations of simulations and other computer-based analytic techniques where machine to machine interoperability is critical. Next steps involve working with subject-matter experts employing digital twins, identifying use cases to test our representations, and clarifying verbal disputes to promote semantic interoperability.